\documentclass[10pt,twocolumn,letterpaper]{article}

\usepackage{cvpr}              % To produce the CAMERA-READY version
\newcommand{\ignore}[1]{}

\newcommand{\ba}{\begin{array}}
        \newcommand{\ea}{\end{array}}
\newcommand{\bc}{\begin{center}}
        \newcommand{\ec}{\end{center}}
\newcommand{\be}{\begin{enumerate}}
        \newcommand{\ee}{\end{enumerate}}
\newcommand{\bea}{\begin{eqnarray}}
        \newcommand{\eea}{\end{eqnarray}}
\newcommand{\beas}{\begin{eqnarray*}}
        \newcommand{\eeas}{\end{eqnarray*}}
\newcommand{\beq}{\begin{equation}}
        \newcommand{\eeq}{\end{equation}}
\newcommand{\bfig}{\begin{figure}}
        \newcommand{\efig}{\end{figure}}
\newcommand{\bi}{\begin{itemize}}
        \newcommand{\ei}{\end{itemize}}
\newcommand{\bpic}{\begin{picture}}
        \newcommand{\epic}{\end{picture}}
\newcommand{\btabular}{\begin{tabular}}
        \newcommand{\etabular}{\end{tabular}}
\newcommand{\btable}{\begin{table}}
        \newcommand{\etable}{\end{table}}

\newcommand{\es}{\vfill
    \rule[-6mm]{170mm}{0.7mm} \\
    \redw{{\tiny
                \hfill S-\theslide}}
    \end{slide}}

\newcommand{\matxx}[1]{{\mathbf #1}}

\def \hbar {{\bar{h}}}

\def \matC {{\matxx{C}}}

\def \matT {{\matxx{T}}}

\def \matX {{\matxx{X}}}

\makeatletter
\renewcommand*\env@matrix[1][*\c@MaxMatrixCols c]{%
    \hskip -\arraycolsep
    \let\@ifnextchar\new@ifnextchar
    \array{#1}}
\makeatother

\newcommand{\RR}{\mathbb{R}}

\newcommand{\Sim}[1]{\ensuremath{\mathbf{Sim}(#1)}}

\font\Bigmath=cmsy10 scaled \magstep2
\def\dplus{\mathrel{%
        \ooalign{$+$\cr\hss\lower.255ex\hbox{\Bigmath\char5}\hss}}}
\def\dminus{\mathrel{%
        \ooalign{$-$\cr\hss\lower.255ex\hbox{\Bigmath\char5}\hss}}}

\def\bea#1\eea{\begin{align}#1\end{align}}
\def\beas#1\eeas{\begin{align*}#1\end{align*}}

\usepackage{microtype}
\newcommand{\method}{\textbf{Online3R} }

\usepackage{xspace}

\usepackage{multirow} 
\usepackage{boldline}

\newcommand{\mcal}[1]{\mathcal{#1}}

\newcommand{\I}{\mcal{I}}

\newcommand{\It}{\I^t}

\newcommand{\X}{\matX}

\newcommand{\Xtt}{\X^t_{t}}
\newcommand{\Xlt}{\X^l_{t}}
\newcommand{\Xlonel}{\X^l_{l-1}}
\newcommand{\Xllone}{\X^{l1}_{l}}
\newcommand{\Xlltwo}{\X^{l2}_{l}}
\newcommand{\Xlhone}{\X^{h1}_{l}}
\newcommand{\Xlhtwo}{\X^{h2}_{l}}
\newcommand{\Xpre}{\X^{l-1}_{l-1}}

\newcommand{\Xcanon}{\tilde{\matX}}

\newcommand{\Xcanonll}{\Xcanon^l_{l}}

\newcommand{\Xcanonew}{\Xcanon^{l-1}_{l-1}}

\newcommand{\Ccanon}{\tilde{\matC}}

\newcommand{\Ccanonll}{\Ccanon^l_{l}}

\newcommand{\C}{\matC}

\newcommand{\Clt}{\C^l_t}

\newcommand{\Tlt}{\matT_{lt}}

\definecolor{cvprblue}{rgb}{0.21,0.49,0.74}
\usepackage[pagebackref,breaklinks,colorlinks,allcolors=cvprblue]{hyperref}
\usepackage{multirow}
\usepackage{booktabs}
\usepackage{tabularx}
\usepackage{amsmath}
\usepackage{amssymb}
\usepackage{graphicx}
\usepackage{array}
\usepackage{algorithm}
\usepackage{amsmath}
\usepackage{amsfonts}
\usepackage{algpseudocode}
\usepackage{xcolor} % For \textcolor
\usepackage{hyperref}
\usepackage[accsupp]{axessibility}

\def\paperID{} % *** Enter the Paper ID here
\def\confName{CVPR}
\def\confYear{2026}

\title{Online3R: Online Learning for Consistent Sequential Reconstruction Based on Geometry Foundation Model}

\author{
    Shunkai Zhou\textsuperscript{1, 2} \quad
    Zike Yan\textsuperscript{3} \quad
    Fei Xue\textsuperscript{4} \quad
    Dong Wu\textsuperscript{1, 2} \quad
    Yuchen Deng\textsuperscript{5} \quad
    Hongbin Zha\textsuperscript{1, 2, 6, \dag} \\
    \\
    \textsuperscript{1}School of Intelligence Science and Technology, Peking University \\
    \textsuperscript{2}State Key Laboratory of General Artificial Intelligence \\
    \textsuperscript{3}T-Stone Robotics Institute, The Chinese University of Hong Kong \\
    \textsuperscript{4}NVIDIA \quad \textsuperscript{5}College of Computer and Information Science, Southwest University \\
    \textsuperscript{6}School of Artificial Intelligence and Computer Science, Anqing Normal University
}

\begin{document}
\maketitle
\begin{abstract}

We present Online3R, a new sequential reconstruction framework that is capable of adapting to new scenes through online learning, effectively resolving inconsistency issues.
Specifically, we introduce a set of learnable lightweight visual prompts into a pretrained, frozen geometry foundation model to capture the knowledge of new environments while preserving the fundamental capability of the foundation model for geometry prediction. To solve the problems of missing groundtruth and the requirement of high efficiency when updating these visual prompts at test time, we introduce a local-global self-supervised learning strategy by enforcing the local and global consistency constraints on predictions. The local consistency constraints are conducted on intermediate and previously local fused results, enabling the model to be trained with high-quality pseudo groundtruth signals; the global consistency constraints are operated on sparse keyframes spanning long distances rather than per frame, allowing the model to learn from a consistent prediction over a long trajectory in an efficient way. Our experiments demonstrate that Online3R outperforms previous state-of-the-art methods on various benchmarks. Project page: \url{https://shunkaizhou.github.io/online3r-1.0/}
\end{abstract}    
\section{Introduction}
\label{sec:intro}

Recovering consistent geometry from multi-view images is a key technique to various applications such as Robotics, Virtual Reality and Augmented Reality. This task has been explored for decades from a geometric perspective~\cite{schonberger_structure-from-motion_2016}, but it has suffered from issues of complexity and computational cost. Recently, with the great success of deep learning, many excellent end-to-end geometry foundation models have been proposed to handle two-view~\cite{Wang2024dust3r, Duisterhof2025mast3rsfm} or multiview inputs~\cite{Wang2025vggt,wang2025pi3}. These approaches effectively reduce complexity and computational cost through a large-scale pre-trained feed-forward network.

A recent series of studies has begun to leverage geometry foundational models for sequential reconstruction by training specific models~\cite{Wang2025cut3r, Wang2025spann3r, liu2025slam3r, wu2025point3r} that take video stream data as input or leveraging pretrained foundation models~\cite{Murai2025mast, maggio2025vggt} to provide initial results and then refining them with geometric constraints, but they are still suffering from limited consistency, particularly in completely new scenes. We argue that it is difficult to train a perfect model that is able to work well in all scenes. Instead, the ability to adapt to new environments is necessary.

In this paper, we introduce \method - online learning for consistent sequential reconstruction based on geometry foundation model, a new sequential reconstruction framework that is able to adapt to new environment directly from streaming data at test time. Specifically, \method is built on a pretrained geometry foundation model (\eg, Mast3R~\cite{leroy2024grounding}) to preserve the fundamental ability of recovering coordinate aligned point clouds. In order to enhance the model's capability to adapt to new scenes, we introduce a set of visual prompts~\cite{jia2022visual, yuan2025test3r} to learn the specific scene priors at test time, which are missing in the pretrained sequential reconstruction methods. As these visual prompts are lightweight, they can be easily updated in an online fashion. However, this is not trivial task due to the missing groundtruth at test time and the high efficiency requirement for sequential reconstruction. 

To solve the two aforementioned problems, we introduce a self-supervised learning mechanism that derives local and global consistency constraints from historical results to enhance subsequent predictions. Locally, we dynamically generate pseudo groundtruth by fusing the past outputs of local windows through a confidence-weighted averaging scheme~\cite{Murai2025mast}. Compared with intermediate outputs, the aggregated predictions are more accurate and thus could be used as supervised signals to enforce the local consistency of subsequent predictions. Nevertheless, pure local constraints may lead to overfitting and error accumulation. Therefore, we adopt additional global constraints that force the model's geometric prediction for given keyframes to be invariant to different historical reference frames, allowing the model to maintain a coherent representation of the entire scene over long trajectories, which guarantees the consistency while preserving the efficiency.

Our contributions are summarized as follows:
\begin{itemize}

    \item We introduce \method, a novel online learning-driven sequential reconstruction framework based on a geometry foundation model. It empowers frozen pre-trained models to adapt to novel scenes, effectively ensuring consistency in sequential reconstruction.
    
    \item We propose a local-global self-supervised mechanism to guide this online learning process. We leverage a \textit{local consistency constraint} derived from temporally fused geometry to enhance the accuracy of subsequent predictions, and a \textit{global consistency constraint} that enforces geometric consistency across distant views to mitigate long-term drift, thereby ensuring both global consistency and computational efficiency.
    
    \item We demonstrate through extensive experiments that our parameter-efficient, prompt-based approach achieves state-of-the-art performance on various 3D reconstruction benchmarks, validating the effectiveness of our online learning strategy.
    
\end{itemize}
\section{Related Work}
\label{sec:related_work}

\paragraph{Geometry Foundation Model.}
Conventional 3D reconstruction approaches~\cite{schonberger_structure-from-motion_2016, oliensis2000critique, ozyecsil2017survey} are founded on principles of visual geometry, employing iterative optimization techniques such as Bundle Adjustment (BA) to refine results~\cite{furukawa2015multi, galliani2015massively, schonberger2016pixelwise, wang2023adaptive}. A significant trend has been the shift from traditional optimization-based methods to direct inference using geometry foundation model~\cite{Wang2024dust3r, leroy2024grounding, Duisterhof2025mast3rsfm, Wang2025vggt,wang2025pi3, keetha2025mapanything} to address the issues of complexity and computational cost. These approaches leverage powerful priors learned from large-scale datasets to directly infer 3D scene representations, substantially improving reconstruction efficiency. The pioneering work, DUSt3R~\cite{Wang2024dust3r}, introduces a feed-forward framework capable of directly predicting dense point clouds from image pairs. Built upon DUSt3R, MASt3R~\cite{leroy2024grounding} enhances model robustness by incorporating feature descriptors to establish more reliable cross-image correspondences. More recent works have extended two-view inputs to multiview inputs. VGGT~\cite{Wang2025vggt} employs alternating frame-wise and global-wise attention mechanisms to predict high-quality pointmaps and camera poses from images in a batch. However, due to memory limitations, VGGT~\cite{Wang2025vggt} and its following works~\cite{wang2025pi3} can hardly be directly applied to sequential reconstruction.

\paragraph{Sequential Reconstruction Based on Geometry Foundation Model.}
Some geometry foundation models~\cite{Wang2025spann3r, Wang2025cut3r, yuan2025test3r, Murai2025mast} are specifically designed for sequence processing. Spann3R~\cite{Wang2025spann3r} leverages memory mechanisms to integrate temporal information. CUT3R~\cite{Wang2025cut3r} employs recurrent networks for the same purpose. While highly efficient, these methods can be susceptible to drift over long trajectories as they often lack a backend optimization mechanism to enforce global consistency. Other works, such as MASt3R-SLAM~\cite{Murai2025mast} and VGGT-SLAM~\cite{maggio2025vggt}, utilize predictions from previous foundation models (\eg, MASt3R~\cite{Duisterhof2025mast3rsfm}, VGGT~\cite{Wang2025vggt}) as initial results and perform post-processing to reduce error accumulation. Benefitting from priors of foundation models trained on massive training data, they can produce promising performance in scenes similar to those in the training datasets. However, their accuracy is limited in completely new scenes because of their frozen models. Essentially different from these approaches, our method focuses mainly on enhancing the ability to adapt to new environments by efficiently tuning additional visual prompts.

\begin{figure*}[ht]
\centering
\includegraphics[width=\linewidth]{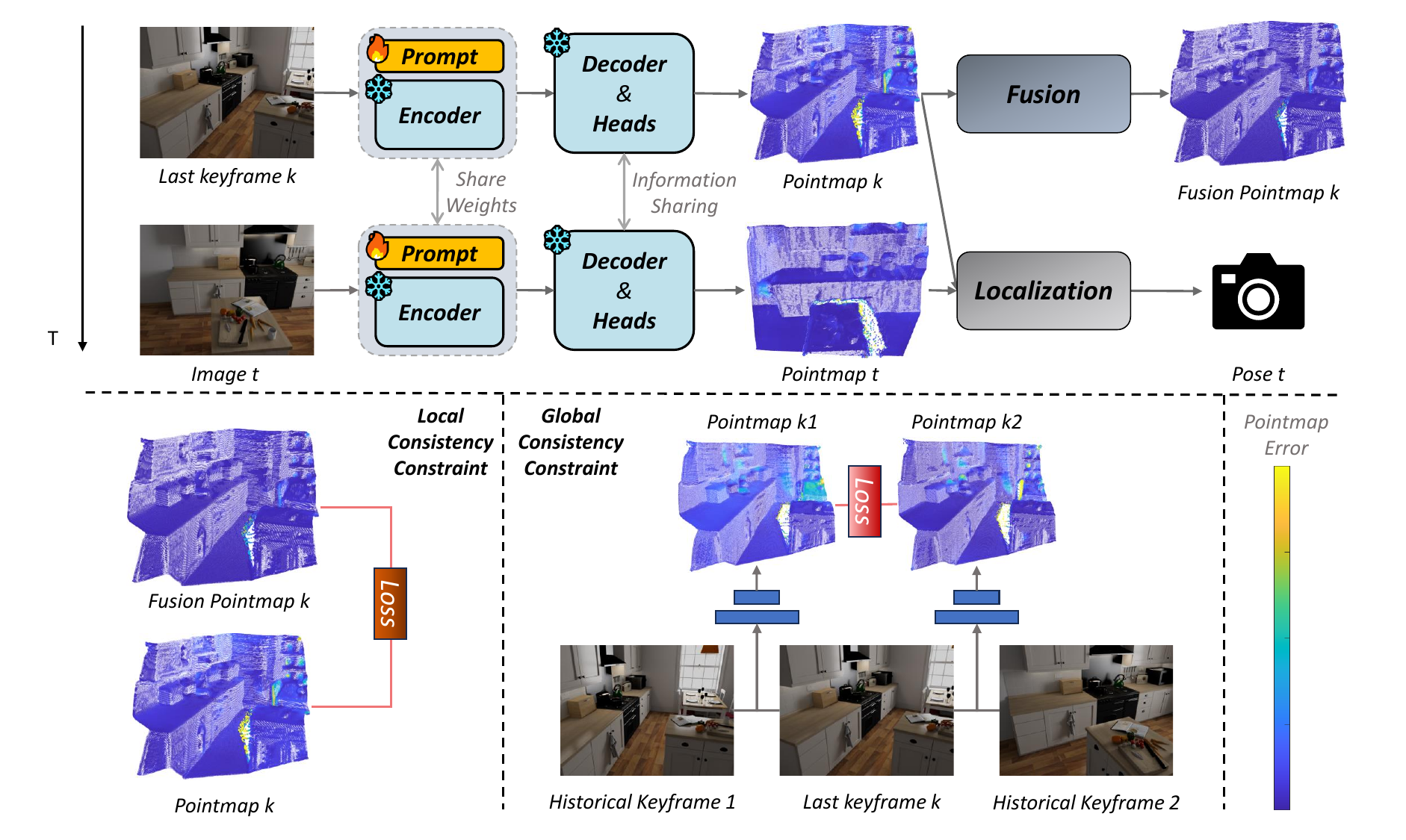}
\caption{\textbf{Overview of our proposed Online3R.} The core of our Online3R lies in constructing self-supervised methods and online prompt tuning, enabling the model to adapt to the current scene and ensuring consistent reconstruction results. We leverage a \textit{local consistency loss} derived from temporally fused geometry to enhance the accuracy of subsequent predictions, and a \textit{global consistency loss} that enforces geometric invariance across distant views to mitigate long-term drift, guaranteeing both consistency and efficiency.}
\label{fig:system_diagram}
\end{figure*}

\paragraph{Parameter-Efficient Network Adaptation.}
For real-world applications requiring high efficiency, full fine-tuning of massive models is computationally expensive. This challenge has driven the development of parameter-efficient fine-tuning (PEFT) techniques. Methods such as Adapters~\cite{pfeiffer-etal-2020-adapterhub}, which insert lightweight modules within Transformer layers, and Low-Rank Adaptation (LoRA)~\cite{hu2022lora}, which optimizes trainable rank-decomposition matrices injected into the attention layers, significantly reduce the number of trained parameters. While highly efficient, these approaches still alter the internal architecture or weights of the original model. A distinct paradigm is prompt tuning~\cite{lester-etal-2021-power}. This technique has been successfully applied to computer vision through Visual Prompt Tuning (VPT)~\cite{jia2022visual}, which introduces a small number of trainable parameters (the ``prompts'') into the input space while keeping the pre-trained backbone completely frozen. These learnable prompts are appended to the input, acting as task-specific instructions that guide the frozen model toward the desired output without modifying its core knowledge. VPT~\cite{jia2022visual} demonstrated that this simple mechanism is remarkably effective, often matching or exceeding the performance of full fine-tuning across various tasks while updating less than 1\% of the parameters.
Recently, efforts have been made to adapt 3D foundation models to specific environments. For instance, LoRA3D~\cite{lu2024lora3d} applies low-rank updates to internal attention weights for scene-specific calibration. In contrast, Test3R~\cite{yuan2025test3r} introduced the prompt tuning mechanism to geometry foundation models, effectively enhancing consistency for offline reconstruction. Specifically, it operates by dividing the input set into multiple triplets and enforcing self-consistency—optimizing the prompts so that the geometry predicted for one frame remains consistent with the reconstruction derived from the other two frames in the triplet. However, this training strategy results in an exponential computational cost, making it unsuitable for sequential reconstruction tasks that demand high efficiency.

\section{Method}
We draw inspiration from VPT~\cite{jia2022visual} and Test3R~\cite{yuan2025test3r}, introducing learnable visual prompts into a pre-trained, frozen 3D foundation model (\ie, MASt3R~\cite{Murai2025mast}) to resolve inconsistencies in sequential reconstruction. However, the absence of ground-truth data and the strict efficiency requirements for online parameter tuning present significant challenges. Therefore, we propose a local-global self-supervised online learning strategy by incorporating local and global consistency constraints on keyframes, which are selected based on an effective feature-matching threshold.

\subsection{Preliminaries: MASt3R-SLAM}
\label{sec:preliminaries}
Our work builds upon the architecture of MASt3R-SLAM~\cite{Murai2025mast}, a real-time, dense SLAM system that leverages a feed-forward network for geometric reconstruction. 

Specifically, the system processes a continuous monocular RGB video stream frame by frame. Each incoming frame $\I$ is paired with the latest keyframe and fed into the network, which outputs their respective per-pixel 3D pointmaps $\X \in \RR^{H\times W\times3}$ and confidence maps $\C \in \RR^{H\times W\times1}$ for both images. Concurrently, it estimates the camera pose $\matT \in \Sim{3}$ for the new frame.

As a keyframe-based system, MASt3R-SLAM~\cite{Murai2025mast} selects a new keyframe when the number of valid matches between the current frame and the last keyframe drops below a specific threshold. The front-end operates by processing each new frame $\It$ in conjunction with the last keyframe $\mathcal{K}^l$, feeding the image pair $(\It, \mathcal{K}^l)$ into the pre-trained MASt3R network~\cite{leroy2024grounding}. The forward pass of the network, denoted as $f_{\theta}$, produces several outputs; among these, the per-pixel 3D pointmap for the current frame in its own coordinate system, $\Xtt \in \mathbb{R}^{H \times W \times 3}$, is our primary focus. The front-end efficiently estimates the relative pose $\Tlt$ between the keyframe $\mathcal{K}^l$ and the current frame $\It$. This is achieved by optimizing the pose to minimize the 3D alignment error between the keyframe's existing pointmap, $\Xcanonll$, and the newly computed pointmap from the current frame, $\Xtt$, transformed by the candidate pose. This process effectively solves for the transformation $\Tlt$ that best aligns the two point clouds, providing a robust and low-latency pose estimate. Since each tracking step re-estimates the pointmap of the latest keyframe, MASt3R-SLAM~\cite{Murai2025mast} performs a confidence-weighted average of these repeatedly computed pointmaps, thereby progressively improving local consistency (see Sec.~\ref{sec:local_consistency}).

Overall, MASt3R-SLAM~\cite{Murai2025mast} provides an ideal foundation for our research. It fuses a powerful feed-forward reconstruction network with a traditional graph-based optimization back-end, resulting in a highly efficient and robust system. However, a critical limitation remains: the core MASt3R~\cite{leroy2024grounding} network operates with completely \textbf{frozen} parameters. It is incapable of adapting its geometric priors to the specific characteristics of the scene being reconstructed. 

\subsection{Online Prompt Tuning}

\subsubsection{Visual Prompt }
\label{sec:prompt_arch}
To enable the frozen, pre-trained MASt3R~\cite{leroy2024grounding} network with the capacity for online adaptation, we introduce a lightweight, learnable module which we term a \textit{prompt}. The core principle is to modulate the response of the powerful, pre-trained model without altering its underlying weights, thereby ensuring both computational efficiency and the preservation of its generalized knowledge.

Formally, similar to \cite{yuan2025test3r}, we insert a set of learnable prompts $\{\mathbf{P}_{i-1}\}_{i=1}^{N_e}$ within MASt3R's encoder~\cite{leroy2024grounding}, which is composed of $N_e$ standard Vision Transformer (ViT) layers. 
An input image is initially partitioned into fixed-size patches, which are then embedded into D-dimensional tokens denoted as $\mathbf{E_0} =\{\mathbf{e}_{0}^k\in\mathbb{R}^D | k\in \mathbb{N}, 1\leq k\leq N_t\}$, where $N_t$ represents the number of image patch tokens. 
Consequently, an encoder layer $E$ augmented with these visual prompts can be formulated as follows:

\begin{equation}
    [\_,\mathbf{E_i}] = E_i([\mathbf{P_{i-1}}, \mathbf{E_{i-1}}])
\end{equation}

\noindent where $[\cdot, \cdot]$ denotes concatenation. Through the self-attention mechanism in subsequent layers, the learnable prompt tokens interact with the image tokens, effectively modulating the feature extraction process to better suit the specific geometric characteristics of the current scene. The output embeddings corresponding to the prompt tokens are discarded.

Consequently, the output pointmaps are no longer solely a function of the input image pair, but are now conditioned on the current state of our prompt, which we denote as $\mathbf{P}_t$ at time $t$. Crucially, this prompt is not static; it is continuously optimized as the system processes the online video stream. When a new frame $I^t$ is paired with the latest keyframe $\mathcal{K}^l$, the prompt-modulated forward pass is represented as:
\begin{equation}
    (\Xtt, \Xlt) = f_{\theta}(\It, \mathcal{K}^l; \mathbf{P}_t).
    \label{eq:prompt_cond}
\end{equation}
\noindent To optimize this prompt over time, our method relies on suitable online supervisory signals, specifically comprising a local consistency constraint and a global consistency constraint.

\subsubsection{Local Consistency Constraint}
\label{sec:local_consistency}

Our local consistency constraint originates from the incremental geometry fusion process inherent to MASt3R-SLAM~\cite{Murai2025mast}. The system does not discard the geometric information from past frames; instead, it continuously refines the pointmap of a keyframe by integrating measurements from subsequent views. As described in~\cite{Murai2025mast}, once the relative pose $\Tlt$ between a new frame $\It$ and the last keyframe $\mathcal{K}^l$ is estimated, the keyframe's canonical pointmap, $\Xcanonll$, is updated via a running weighted average filter applied in a per-pixel manner:

\begin{equation}
\Xcanonll \leftarrow \frac{\Ccanonll \Xcanonll +  \Clt \left(\Tlt \Xlt\right)}{ \Ccanonll + \Clt}, 
\Ccanonll \leftarrow \Ccanonll + \Clt~.
\label{eq:fusion}
\end{equation}
where $\Ccanonll$ is the fused confidence map. This filtering process merges information from multiple viewpoints over time, mitigating noise and single-view ambiguities. The resulting fused pointmap $\Xcanonll$ is a more accurate and geometrically consistent representation of the scene than any single-shot network prediction. We leverage this temporally enhanced geometry as a high-quality \textbf{pseudo ground truth} for our online learning.

The optimization of our prompt $\mathbf{P}$ is triggered whenever a new frame $\It$ is designated as a keyframe, which becomes the new last keyframe $\mathcal{K}^l$. At this moment, the pointmap of the previous keyframe $\mathcal{K}^{l-1}$ is fully updated by the fusion process. We additionally perform a forward pass using the image pair of $\mathcal{K}^{l}$ and $\mathcal{K}^{l-1}$ with the current prompt $\mathbf{P}_{t}$. This pass, $f_{\theta}(\mathcal{K}^{l-1}, \mathcal{K}^{l}; \mathbf{P}_{t})$, yields a direct, single-shot prediction for $\mathcal{K}^{l-1}$'s geometry, denoted as $\Xpre$.

We formulate our local consistency loss, $\mathcal{L}_{\text{local}}$, as the $\ell_1$ distance between the fused pseudo ground truth and the network's direct prediction:
\begin{equation}
    \mathcal{L}_{\text{local}}(\Xcanonew, \Xpre) = \sum_{z}  \left\| \Xcanonew(z) - \Xpre(z) \right\|_1
    \label{eq:local_loss}
\end{equation}
where $z$ iterates over all pixel coordinates. By backpropagating this loss to update only the parameters of the prompt $\mathbf{P}_{t}$, we effectively distill the knowledge from the multi-view fusion back into the feed-forward network, forcing it to produce reconstructions that are more consistent with the temporally aggregated geometry of the scene.

\subsubsection{Global Consistency Constraint}
\label{sec:global_consistency}
Although relying solely on the local consistency loss $\mathcal{L}_{\text{local}}$ improves prediction consistency, it may also lead the model to overfit to the most recent geometric features, causing it to gradually ``forget'' the global structure of the scene.
% Although relying solely on the local consistency loss $\mathcal{L}_{\text{local}}$ improve prediction consistency, also may lead to the model overfitting to the most recent geometric features, causing it to gradually ``forget'' the global structure of the scene. 
To mitigate this temporal drift and enforce a more holistic understanding of the environment, we introduce a global consistency constraint.

This constraint is designed to ensure that the network's prediction for a given keyframe's geometry remains consistent regardless of which historical view is used as a reference. Specifically, when a new keyframe $\mathcal{K}^l$ is created at time $t$, we randomly sample two distinct historical keyframes, $\mathcal{K}^{h1}$ and $\mathcal{K}^{h2}$, from the pose graph, where $h1, h2 < l$. We then perform two independent forward passes through the prompt-modulated network:
\begin{enumerate}
    \item $f_{\theta}(\mathcal{K}^l, \mathcal{K}^{h1}; \mathbf{P}_{t})$, which yields a pointmap prediction $\Xllone$ for the current keyframe.
    \item $f_{\theta}(\mathcal{K}^l, \mathcal{K}^{h2}; \mathbf{P}_{t})$, which yields a second pointmap prediction $\Xlltwo$ for the same keyframe.
\end{enumerate}

\noindent Ideally, these two outputs, $\Xllone$ and $\Xlltwo$, should be identical as they represent the exact same physical geometry. Any deviation indicates a failure to maintain a consistent representation across different parts of the map.
We therefore formulate a global consistency loss, $\mathcal{L}_{\text{global}}$, to penalize this inconsistency. The loss is defined as the $\ell_1$ distance conditioned on two different keyframes:
\begin{equation}
    \mathcal{L}_{\text{global}}(\Xllone, \Xlltwo) = \sum_{z} \left\| \Xllone(z) - \Xlltwo(z) \right\|_1
    \label{eq:global_loss}
\end{equation}
This loss encourages the prompt to learn a representation that is robust to the choice of reference frame, thereby enforcing long-term geometric consistency.
The final objective function for updating our prompt combines both the local and global consistency constraints. The total loss is a weighted sum of the two terms:
\begin{equation}
    \mathcal{L}_{\text{total}} = \lambda \mathcal{L}_{\text{local}} + (1 - \lambda) \mathcal{L}_{\text{global}}
    \label{eq:total_loss}
\end{equation}
where $\lambda$ is a hyperparameter that balances the influence of the two objectives.

\subsection{Online Optimization and Implementation}
\label{sec:implementation}

% Our prompt-driven online learning is an event-based process, seamlessly integrated into the MASt3R-SLAM pipeline. The optimization of the prompt is not performed on every frame, but is instead triggered only upon the creation of a new keyframe. 

Our online prompt tuning strategy is closely related to the emergence of keyframes: whenever a new keyframe appears, we compute both the local and global consistency losses to update the prompt. The local consistency loss is calculated within the local keyframe window, while global consistency loss is computed by randomly sampling two frames from the keyframe buffer. This strategy ensures computational efficiency while allowing the model to adapt at critical moments when significant new information about the scene becomes available. The complete online optimization procedure is summarized in Algorithm~\ref{alg:online_learning}.

\begin{algorithm}[ht]
    \caption{Online Prompt Tuning}
    \renewcommand{\algorithmicrequire}{\textbf{Input:}}
	\renewcommand{\algorithmicensure}{\textbf{Output:}}
    \label{alg:online_learning}
    \begin{algorithmic}[1]
        \Require Image stream $\{\It\}$, $f_{\theta}$

        % \Ensure None
        
        \State $\mathbf{P}_0 \leftarrow \mathbf{0}$  \Comment{Zero-initialize prompt}
        \State $\mathcal{K}^0 \leftarrow \mathcal{I}^0$  \Comment{Initialize the first frame as a keyframe}
        \State $l \leftarrow 1$  \Comment{Keyframe index}
        \For{$t$ in $[1,2,3,\dots]$}
            % \State Process $\It$ with MASt3R-SLAM front-end to track pose $\Tlt$
            % \State $(\mathbf{X}_{l-1}^{l-1}, \mathbf{X}_{l-1}^{t}) \leftarrow f_\theta(\mathcal{K}^{l-1}, \mathcal{I}^t; \mathbf{P}_{t})$
            \State $(\mathbf{X}_{t}^{t}, \mathbf{X}_{t}^{l}) \leftarrow f_\theta(\mathcal{I}^t, \mathcal{K}^{l}; \mathbf{P}_{t})$
            \State $\Xcanonll \leftarrow \texttt{fusion}(\Xcanonll, \mathbf{X}_{t}^{l})$  \Comment{Eq.~\ref{eq:fusion}}
            % \If{$\It$ is designated as a new keyframe}
            \If{\texttt{is\_keyframe}($\mathcal{I}^t$)}
                \State $l \leftarrow l + 1$
                \State $\mathcal{K}^l \leftarrow \mathcal{I}^{t}$
                \State \textcolor{gray}{\# --- Local Consistency Update ---}
                % \State Get fusion pointmap $\Xcanonew$ of previous keyframe $\mathcal{K}^{l-1}$
                \State $(\Xpre, \Xlonel) = f_{\theta}(\mathcal{K}^{l-1}, \mathcal{K}^{l}; \mathbf{P}_t)$
                % \State Compute $\mathcal{L}_{\text{local}}$ using Eq.~\ref{eq:local_loss}
                \State $\mathcal{L}_{\text{local}} \leftarrow \mathcal{L}_{\text{local}}(\Xcanonew, \mathbf{X}_{l-1}^{l-1})$  \Comment{Eq.~\ref{eq:local_loss}}
                \State \textcolor{gray}{\# --- Global Consistency Update ---}
                % \State Sample two distinct historical keyframes $\mathcal{K}^{h1}, \mathcal{K}^{h2}$ from the pose graph
                \State $\mathcal{K}^{h1}, \mathcal{K}^{h2} \leftarrow \texttt{sample}(\{\mathcal{K}\})$ \Comment{sample two previous keyframes}
                \State $(\Xllone, \Xlhone) \leftarrow f_{\theta}(\mathcal{K}^l, \mathcal{K}^{h1}; \mathbf{P}_t)$
                \State $(\Xlltwo, \Xlhtwo) \leftarrow f_{\theta}(\mathcal{K}^l, \mathcal{K}^{h2}; \mathbf{P}_t)$
                % \State Compute $\mathcal{L}_{\text{global}}$ using Eq.~\ref{eq:global_loss}
                \State $\mathcal{L}_{\text{global}} \leftarrow \mathcal{L}_{\text{global}}(\mathbf{X}_{l}^{l1}, \mathbf{X}_{l}^{l2})$ \Comment{Eq.~\ref{eq:global_loss}}
                \State \textcolor{gray}{\# --- Optimizer Step ---}
                \State $\mathcal{L}_{\text{total}} \leftarrow \lambda \mathcal{L}_{\text{local}} + (1 - \lambda) \mathcal{L}_{\text{global}}$ \Comment{Eq.~\ref{eq:total_loss}}
                % \State Backpropagate gradient $\nabla_{\mathbf{P}} \mathcal{L}_{\text{total}}$
                % \State Update prompt: $\mathbf{P}_{t} \leftarrow \text{OptimizerStep}(\mathbf{P}_{t-1}, \nabla_{\mathbf{P}} \mathcal{L}_{\text{total}})$
                \State $\mathbf{P}_t \leftarrow \texttt{optimizer}(\mathbf{P}_t, \nabla_{\mathbf{P}} \mathcal{L}_{\text{total}})$
            \EndIf
            \State $\mathbf{P}_{t+1} \leftarrow \mathbf{P}_t$
        \EndFor
    \end{algorithmic}
\end{algorithm}

\paragraph{Implementation Details.}
Our implementation is built upon the official public release of MASt3R-SLAM. For the learnable prompt, we use $N_p=32$ length, each with a dimension of $D=1024$, matching the feature dimension of the MASt3R encoder. For the optimization, we use the AdamW optimizer with a learning rate of $1 \times 10^{-4}$. The balancing hyperparameter $\lambda$ in our total loss function (Eq.~\ref{eq:total_loss}) is set to $0.5$ across all experiments. For the global consistency constraint, the two historical keyframes are sampled randomly from all previously established keyframes. All experiments are conducted on a single NVIDIA A100 GPU.
\section{Experiments}

\begin{table*}[t]
\centering
\caption{\textbf{Absolute trajectory error (ATE (m)$\downarrow$) on TUM RGB-D \cite{sturm_benchmark_2012}.} We evaluate methods with and without ground-truth camera intrinsics as inputs (noted as ``Calibrated'' and ``Uncalibrated''). The \textbf{best} and the \underline{second best} results of each type are highlighted.}\label{tab:tum_ate}
% \scriptsized
\scalebox{0.925}{
\begin{tabular}{l|l|ccccccccc|c} %\toprule
\toprule
\textbf{Type} & \textbf{Methods} &\textbf{360} &\textbf{desk} &\textbf{desk2} &\textbf{floor} &\textbf{plant} &\textbf{room } &\textbf{rpy} &\textbf{teddy} &\textbf{xyz} &\textbf{avg} \\
\midrule
\multirow{8}{*}{Calibrated} & ORB-SLAM3~\cite{campos_orbslam3_2021} & -- &\underline{0.017} &0.210 & -- &0.034 & -- & -- & -- &\underline{0.009} & -- \\
&DeepV2D~\cite{teed_deepv2d_2020} &0.243 &0.166 &0.379 &1.653 &0.203 &0.246 &0.105 &0.316 &0.064 &0.375 \\
&DeepFactors \cite{czarnowski_deepfactors_2020} &0.159 &0.170 &0.253 &0.169 &0.305 &0.364 &0.043 &0.601 &0.035 &0.233 \\
&DPV-SLAM \cite{lipson2024deep} &0.112 &0.018 &0.029 &0.057 &0.021 &0.330 &0.030 &0.084 &0.010 &0.076 \\
&DPV-SLAM++ \cite{lipson2024deep} &0.132 &0.018 &0.029 &0.050 &0.022 &0.096 &0.032 &0.098 &0.010 &0.054 \\
&GO-SLAM \cite{zhang2023goslam} &0.089 &\textbf{0.016} &\underline{0.028} &\underline{0.025} &0.026 &\underline{0.052} &\textbf{0.019} &0.048 &0.010 &0.035 \\
&DROID-SLAM \cite{teed_droid_2021} &0.111 &0.018 &0.042 &\textbf{0.021} &\textbf{0.016} &\textbf{0.049} &0.026 &0.048 &0.012 &0.038 \\
&MASt3R-SLAM~\cite{Murai2025mast} &\underline{0.049} &\textbf{0.016} &\underline{0.024} &\underline{0.025} &0.020 &0.061 &0.027 &\underline{0.041} &\underline{0.009} &\underline{0.030} \\
&\textbf{Ours} &\textbf{0.044} &\textbf{0.016} &\textbf{0.022} &0.026 &\underline{0.017} &0.054 &\underline{0.023} &\textbf{0.035} &\textbf{0.008} &\textbf{0.027} \\
\midrule
\multirow{6}{*}{Uncalibrated} &DROID-SLAM* \cite{teed_droid_2021, veicht2024geocalib} &0.202 &0.032 &0.091 &0.064 &0.045 &0.918 &0.056 &\underline{0.045} &\textbf{0.012} &0.158 \\
&Spann3R~\cite{Wang2025spann3r} &0.146 &0.191 &0.199 &0.364 &0.334 &0.526 &0.050 &0.248 &0.091 &0.238 \\
&CUT3R~\cite{Wang2025cut3r} &0.122 &0.045 &0.089 &\textbf{0.047} &0.057 &\textbf{0.073} &\textbf{0.029} &\textbf{0.037} &0.023 &\underline{0.058} \\
&Point3R~\cite{wu2025point3r} &0.138 &0.114 &0.179 &0.092 &0.098 &\underline{0.110} &0.043 &0.090 &0.046 &0.101 \\
&MASt3R-SLAM*~\cite{Murai2025mast} &\underline{0.070} &\underline{0.035} &\textbf{0.055} &0.056 &\underline{0.035} &0.118 &0.041 &0.114 &\underline{0.020} &0.060 \\
&\textbf{Ours*} &\textbf{0.065} &\textbf{0.034} &\underline{0.056} &\underline{0.055} &\textbf{0.032} & \underline{0.110} & \underline{0.040} &0.095 &\underline{0.020} &\textbf{0.056} \\
\bottomrule
\end{tabular}
}
\end{table*}

\paragraph{Experimental Overview.}
Our method addresses the challenge of consistent online reconstruction by tackling its two core dependencies—accurate camera pose estimation and high-fidelity per-frame geometry prediction—both of which are essential for forming a globally coherent map. Therefore, we structure our quantitative experiments to evaluate these two aspects separately: Camera Pose Estimation (in Sec.~\ref{sec:4.1}) and Dense Geometry Evaluation (in Sec.~\ref{sec:4.2}). 
To visually demonstrate the overall impact on consistency, our qualitative analysis (in Sec.~\ref{sec:4.2}) compares the final reconstructed point clouds from different methods, as this serves as a comprehensive indicator of the combined performance of both pose and geometry estimation.
Furthermore, computational efficiency is a critical requirement for online tasks. While our online learning strategy is designed to enhance reconstruction consistency, it introduces a marginal computational overhead. We therefore conduct an Efficiency Analysis (in Sec.~\ref{sec:4.4}) to evaluate this trade-off. Finally, to validate our design choices and ablate the contribution of each key element, we present a Component Analysis (in Sec.~\ref{sec:4.3}).

\paragraph{Benchmark.}
We demonstrate the effectiveness of our Online3R on multiple benchmarks: for camera pose, we evaluate on both the TUM RGB-D~\cite{sturm_benchmark_2012} and NRGBD~\cite{azinovic2022neural} datasets; for dense geometry, we follow Point3R~\cite{wu2025point3r} evaluating on both the 7-Scenes \cite{shotton_7scenes_2013} and NRGBD~\cite{azinovic2022neural} datasets. It is important to note that all experiments are performed under the monocular RGB setting, without using depth information as input. These datasets provide a diverse set of indoor environments and camera trajectories, forming a robust testbed for our method.

\paragraph{Experimental Setup.}
All our experiments are executed on a server equipped with an INTEL (R) XEON (R) PLATINUM 8558P CPU and a single NVIDIA A100 GPU. Our system, which incorporates the online prompt updating mechanism, runs at approximately 10 FPS.  Note that similar to MASt3R-SLAM~\cite{Murai2025mast} our method utilizes the full-resolution outputs from the foundational MASt3R~\cite{leroy2024grounding} network, which resizes the longest image dimension to 512 pixels.

\subsection{Camera Pose Estimation}
\label{sec:4.1}
We report the Root Mean Square Error (RMSE) of the Absolute Trajectory Error (ATE) in meters for all datasets. Due to the inherent lack of scale information in monocular reconstruction, we align the estimated trajectory with the ground truth. Geometry foundation model based methods typically do not require intrinsic calibration, whereas traditional methods often do. Therefore, we compare the results in two modes: calibrated and uncalibrated. For methods supporting both modes, we use * to denote the uncalibrated.

\setlength{\tabcolsep}{2pt}
\begin{table}[t]
\centering
\caption{\textbf{Absolute trajectory error (ATE (m)$\downarrow$) on NRGBD \cite{azinovic2022neural}.} 
MASt3R-SLAM*~\cite{Murai2025mast} is abbreviated as M-S*. The \textbf{best} and the \underline{second best} results are highlighted.}\label{tab:nrgbd_pose}
% \scriptsize
\scalebox{0.765}{
\begin{tabular}{@{}l|ccccccccc|c@{}} %\toprule
\toprule
Methods        & brea  & comp  & gree  & grey  & kitc  & morn  & stai  & thin  & whi   & avg   \\
\midrule
Spann3R        & 1.464 & 1.980 & 0.981 & 1.269 & 1.989 & 0.631 & 1.997 & 0.368 & 2.319 & 1.444 \\
CUT3R          & 0.566 & 1.450 & 0.868 & 0.922 & 1.099 & 0.329 & 0.466 & 0.204 & 1.847 & 0.861 \\
Point3R        & 0.416 & 1.155 & 0.302 & 0.659 & 1.354 & 0.269 & 0.657 & 0.133 & 0.593 & 0.615 \\
M-S*           & \underline{0.109} & \underline{0.079} & \underline{0.095} & \textbf{0.089} & \underline{0.176} & \underline{0.058} & \underline{0.056} & \textbf{0.046} & \underline{0.105} & \underline{0.090} \\
\textbf{Ours*} & \textbf{0.102} & \textbf{0.063} & \textbf{0.084} & \underline{0.102} & \textbf{0.118} & \textbf{0.057} & \textbf{0.047} & \underline{0.047} & \textbf{0.066} & \textbf{0.076} \\
\bottomrule
\end{tabular}
}
\end{table}

\paragraph{TUM RGB-D.} As shown in Table~\ref{tab:tum_ate}, our Online3R achieves state-of-the-art performance on average in both calibrated and uncalibrated modes. In the calibrated scenario, Online3R surpasses the baseline MASt3R-SLAM~\cite{Murai2025mast} and other leading methods like DROID-SLAM~\cite{teed_droid_2021}. This demonstrates that our online prompt tuning mechanism effectively improves the consistence of 3D prior, leading to more accurate pose estimation. In the more challenging uncalibrated setting, Ours* improves upon the MASt3R-SLAM*~\cite{Murai2025mast} baseline and demonstrates certain improvements over Spann3R~\cite{Wang2025spann3r}, CUT3R~\cite{Wang2025cut3r} and Point3R~\cite{wu2025point3r}.

\begin{table*}[t]
  \centering
  % \caption{\textbf{Quantitative 3D reconstruction results on 7-scenes and NRGBD datasets.} We use ``GA'' to mark methods with global alignment, and use ``Optim.'' and ``Onl.'' to distinguish between optimization-based and online methods~\cite{Wang2025cut3r}. Our method achieves competitive or better performance than those optimization-based methods and current online methods.}
  \caption{\textbf{Quantitative 3D reconstruction results on 7-scenes and NRGBD datasets.} We categorize methods into ``Offline'' and ``Online''~\cite{Wang2025cut3r}, and use ``GA'' to mark methods with global alignment. The \textbf{best} and the \underline{second best} results of all methods are highlighted. Our method achieves competitive or better performance than those offline or online methods.}
  \label{tab: 3d}
  % \footnotesize
  \scalebox{0.95}{
  \begin{tabular}{ @{\hskip 1em}  l @{\hskip 1.5em} | @{\hskip 1.5em} l @{\hskip 1.5em} | @{\hskip 1.5em} c @{\hskip 1.5em} c @{\hskip 1.5em} c @{\hskip 1.5em} c @{\hskip 1.5em} c @{\hskip 1.5em} c  @{\hskip 1em} }
    \toprule
    \multirow{2}{*}{\textbf{Types}} & \multirow{2}{*}{\textbf{Methods}} &\multicolumn{3}{c}{\textbf{7-scenes}} & \multicolumn{3}{c}{\textbf{NRGBD}} \\
    & & {Acc$\downarrow$} & {Comp$\downarrow$} & {Chamf$\downarrow$} & {Acc$\downarrow$} & {Comp$\downarrow$} & {Chamf$\downarrow$} \\
    \midrule
    \multirow{4}{*}{Offline} & DUSt3R-GA~\cite{Wang2024dust3r}      & 0.146 & 0.181 & 0.164 & 0.144 & 0.154 & 0.149 \\
    & MASt3R-GA~\cite{leroy2024grounding}  & 0.185 & 0.180 & 0.183 & 0.085 & \textbf{0.063} & \underline{0.074} \\
    & MonST3R-GA~\cite{zhang2024monst3r}   & 0.248 & 0.266 & 0.257 & 0.272 & 0.287 & 0.280 \\
    & Test3R~\cite{yuan2025test3r}         & 0.105 & 0.136 & 0.121 & 0.083 &  0.079 & 0.081 \\
    \midrule
    \multirow{5}{*}{Online} & Spann3R~\cite{Wang2025spann3r}       & 0.298 & 0.205 & 0.252 & 0.416 & 0.417 & 0.417 \\
    & CUT3R~\cite{Wang2025cut3r}           & 0.126 & 0.154 & 0.140 & 0.099 & 0.076 & 0.088 \\
    & Point3R~\cite{wu2025point3r}         & 0.124 & 0.139 & 0.132 & 0.079 & \underline{0.073} & 0.076 \\
    & MASt3R-SLAM*~\cite{Murai2025mast}    & \underline{0.068} & \textbf{0.045} & \underline{0.056} & \underline{0.065} & 0.094 & 0.080 \\
    & \textbf{Ours*}                       & \textbf{0.039} & \underline{0.067} & \textbf{0.053} & \textbf{0.053} & 0.093 & \textbf{0.073} \\
    \bottomrule
  \end{tabular}
  }
  \label{tab:3d_recon_7s_nr}
  \vspace{-3mm}
\end{table*}

\begin{figure*}[ht]
\centering
\includegraphics[width=\linewidth]{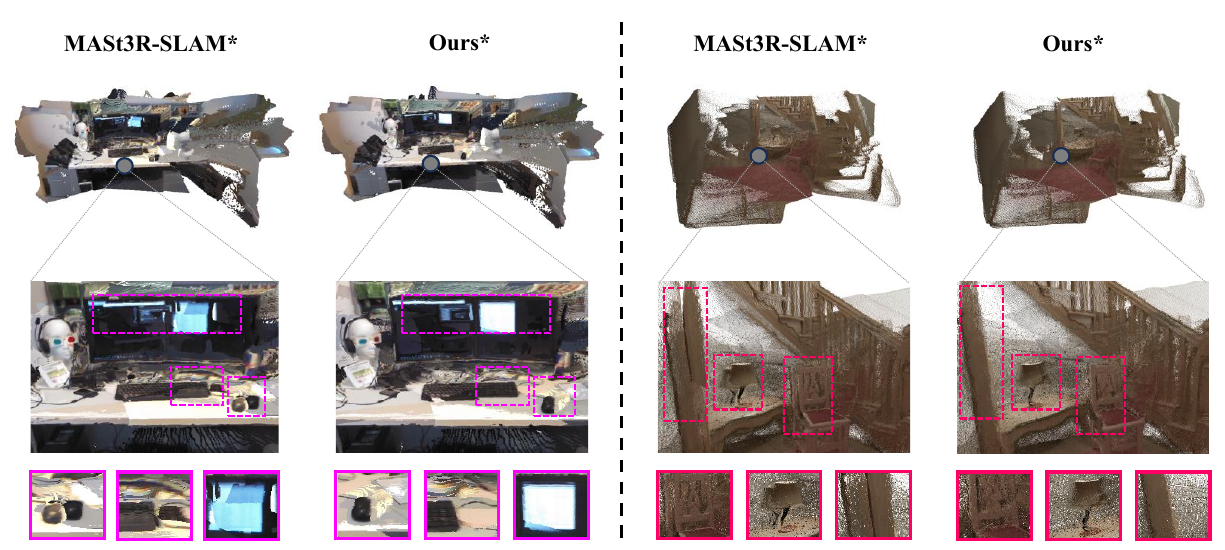}
\caption{\textbf{Qualitative Comparison on 3D Reconstruction Consistency.} We present reconstruction results on two sequences, with \textit{7-Scenes-heads} on the left and \textit{NRGBD-staircase} on the right, separated by dashed lines. The first row shows the global point cloud reconstruction from a far viewpoint. The second row zooms in to view the near viewpoint. The third row highlights details using bounding boxes. It is evident that our method achieves better consistency compared to MASt3R-SLAM*.}
\label{fig:dingxing}
\end{figure*}

\paragraph{NRGBD.} On the NRGBD~\cite{azinovic2022neural} dataset, presented in Table~\ref{tab:nrgbd_pose}, our method shows a significant performance advantage. Note that unlike the offline 3D reconstruction foundation model~\cite{leroy2024grounding} used in our Online3R and MASt3R-SLAM~\cite{Murai2025mast}, Spann3R~\cite{Wang2025spann3r}, CUT3R~\cite{Wang2025cut3r} and Point3R~\cite{wu2025point3r} are designed for online reconstruction. However, constrained by the model's generalization capabilities, their outputs in test scenes often exhibit significant cumulative errors, ultimately leading to pose drift. MASt3R-SLAM~\cite{Murai2025mast} partially mitigates reconstruction inconsistencies through map fusion. Nevertheless, since its pose estimation still relies on the consistency of network outputs, it remains challenging to fully overcome cumulative error. These findings collectively demonstrate that continuous online learning is crucial for 3D reconstruction foundation models.

\begin{figure}[t]
	\centering
	\includegraphics[width=0.98\columnwidth]{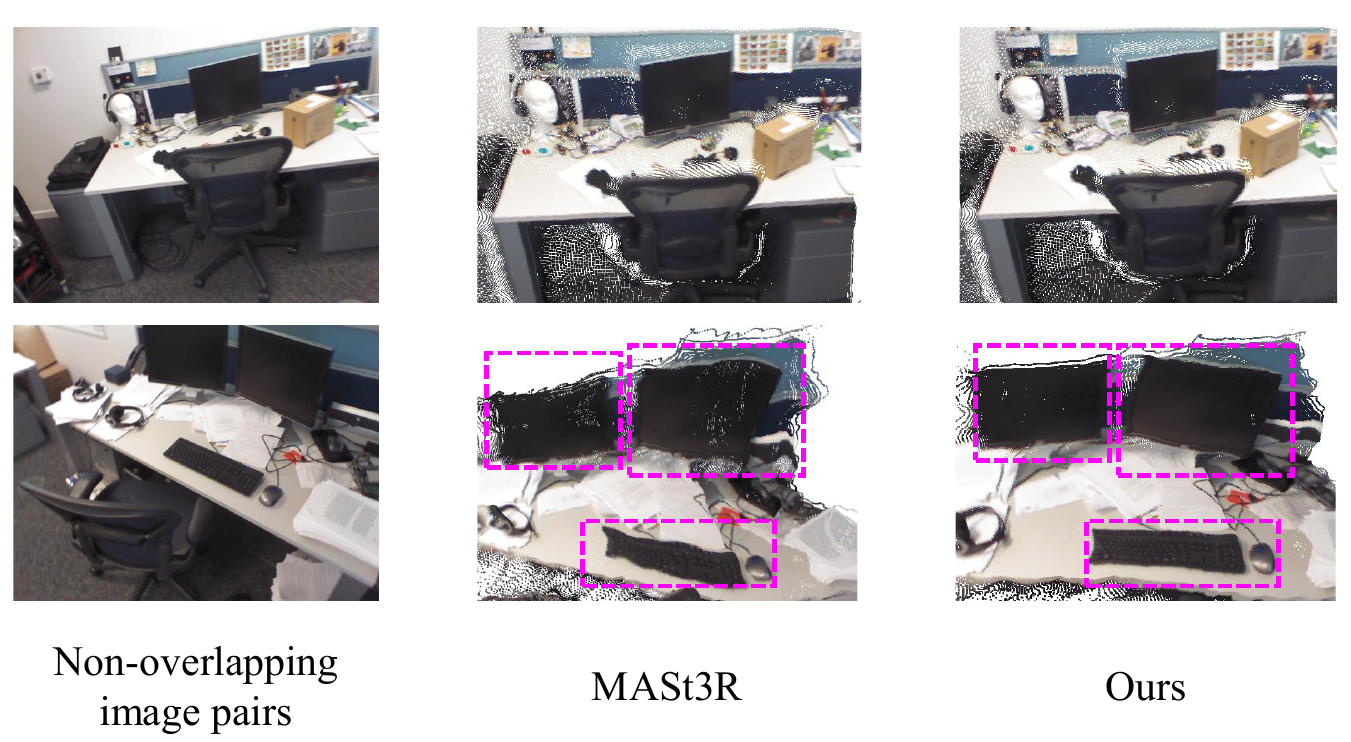}
	\caption{Qualitative results for reconstruction of non-overlapping image pairs.}
	\label{fig:zoom_ablation} 
\end{figure}

\subsection{Dense Geometry Evaluation}
\label{sec:4.2}
We evaluate the quality of the dense geometry generated by our system against other state-of-the-art methods on the 7-Scenes~\cite{shotton_7scenes_2013} and NRGBD~\cite{azinovic2022neural} datasets. All methods are run in the uncalibrated setting to demonstrate robustness to unknown camera intrinsics. For a comprehensive comparison, we report three standard metrics for point cloud similarity: Accuracy, Completion, and Chamfer Distance. Accuracy is the average distance from each point in our estimated reconstruction to its nearest neighbor in the ground truth point cloud, which is the most important metric for evaluating consistency. Conversely, Completion is the average distance from each ground truth point to its nearest neighbor in our estimate, measuring how much of the scene is captured. The Chamfer Distance is the average of Accuracy and Completion. For all metrics, a lower value indicates better performance. We use a maximum distance threshold of 0.5m to discard outliers during calculation.

As shown in Table~\ref{tab: 3d}, our Online3R achieves more accurate geometric results, even surpassing offline reconstruction methods like DUSt3R-GA~\cite{Wang2024dust3r} and MASt3R-GA~\cite{leroy2024grounding}, demonstrating that our online learning strategy effectively ensures reconstruction consistency. This superiority likely stems from two factors: 1. we constructed local consistency constraint based on the results of local pointmap fusion, significantly enhancing local geometric accuracy, 2. our online global consistency supervision is designed to be applied on keyframes, providing broader coverage that enables the prompt to learn comprehensive scene information.

Figure~\ref{fig:dingxing} shows a comparison between our Online3R and MASt3R-SLAM~\cite{Murai2025mast} for dense reconstruction. To ensure a fair comparison, neither method applied confidence-based filtering to the point cloud, resulting in the presence of some outliers. Compared to MASt3R-SLAM~\cite{Murai2025mast}, Online3R maintains consistency in sequential reconstructions, preventing misalignment in reconstructions of the same region.

\subsection{Efficiency Analysis}
\label{sec:4.4}
Since online learning strategies bring additional computational consumption, we compared our FPS with MASt3R-SLAM~\cite{Murai2025mast} across all sequences in NRGBD~\cite{azinovic2022neural}. As shown in Table~\ref{tab:Efficiency}, we achieved limited FPS degradation while demonstrating significant improvements in ATE results, attributable to our lightweight prompt design and efficient online learning strategy.

\subsection{Component Analysis}
\label{sec:4.3}

To evaluate the impact of our proposed constraints, we conducted an ablation study on the 7-scenes~\cite{shotton_7scenes_2013} dataset. As reported in Table~\ref{tab:ablation}, we compare variants of Online3R utilizing exclusively local consistency (Local*), exclusively global consistency (Global*), and their combination (Full*). Measured by average accuracy (Acc) across all sequences, our findings indicate that while each constraint independently enhances geometric accuracy, their joint application yields the optimal performance.

Furthermore, we observe that through online learning, the prompt progressively encodes implicit scene-specific knowledge. As illustrated in Figure~\ref{fig:zoom_ablation}, when presented with non-overlapping views, the baseline MASt3R~\cite{leroy2024grounding} reliably estimates the reference frame but fails to reconstruct the non-reference geometry. We attribute this failure to the foundation model's lack of current 3D scene priors. In contrast, our approach accurately recovers the geometry of the non-reference frame. This validates the efficacy of our online learning strategy and highlights the potential of prompts for implicit 3D scene representation.

\begin{table}[!t]
    \centering
    \setlength{\tabcolsep}{1pt}
    \scalebox{1}{
    \begin{minipage}[t]{0.48\columnwidth}
        \centering
        \caption{Efficiency analysis.}\label{tab:Efficiency}
        \footnotesize
        \renewcommand{\arraystretch}{1.675} 
        \begin{tabular}{ @{\hskip 0.5em} l @{\hskip 0.5em} | @{\hskip 0.5em} c @{\hskip 0.5em} c @{\hskip 0.5em} c @{\hskip 0.5em} } %\toprule
        \toprule
        Methods   & ATE$\downarrow$ & \#iter. & FPS \\
        \midrule
        M-S*      & 0.090 & -     & 13.2 \\
        Ours*     & 0.076 & 32    & 10.0 \\
        \bottomrule
        \end{tabular}
    \end{minipage}%
    \hfill
    \begin{minipage}[t]{0.48\columnwidth}
        \centering
    \caption{Ablation study.}\label{tab:ablation}
    \footnotesize
    \begin{tabular}{@{\hskip 1em}l@{\hskip 1em} |@{\hskip 1em} c@{\hskip 1em}} %\toprule
    \toprule
    Methods  & Acc$\downarrow$ \\
    \midrule
    M-S*     & 0.068 \\
    Local*   & 0.042 \\
    Global*  & 0.044 \\
    Full*    & 0.039 \\
    \bottomrule
    \end{tabular}
    \end{minipage}
    }
\end{table}

\section{Limitations and Future Work}
Our method incorporates the computation of local and global constraints as well as online optimization based on Visual Prompt Tuning, which imposes additional computational overhead on the system. Meanwhile, our current system is only applicable to sequential reconstruction of 3D static scenes. Extending our method to dynamic scenarios, enabling 3D foundation models to continuously adapt to more challenging 4D dynamic scenes, remains an interesting direction that we leave for future work.
\section{Conclusion}
We present Online3R, an online learning framework that resolves the issue of sequential reconstruction inconsistency of geometry foundation models when applied to new scenes. Our key innovation is to adapt a frozen foundation model by online tuning a set of lightweight visual prompts at test-time. This process allows the model to learn a coherent, scene-specific representation that directly addresses geometric inconsistencies in sequential data. The online adaptation is guided by a local-global self-supervised strategy, ensuring efficient learning without ground truth. Experiments demonstrate that Online3R produces substantially more consistent reconstructions than prior state-of-the-art methods.
\section{Acknowledgement}
We gratefully acknowledge the anonymous reviewers and Area Chairs for their valuable comments and suggestions. We also extend our sincere gratitude to Yingdian Cao for his invaluable assistance in manuscript preparation and formatting.
This work is supported by NSFC (U22A2061) and 230601GP0004.

{
    \small
    \bibliographystyle{ieeenat_fullname}
    \bibliography{main}
}

\end{document}